\documentclass[10pt,twocolumn,letterpaper]{article}

\usepackage[pagenumbers]{preprint}

\usepackage{amsmath}
\usepackage{amssymb}
\usepackage{booktabs}
\usepackage{graphicx}
\usepackage{microtype}
\usepackage{multirow}
\usepackage{tabularx}
\usepackage{longtable}
\usepackage{pdflscape}
\usepackage{xcolor}
\usepackage{xspace}

\newcommand{\sourcepath}[1]{\path{#1}}

\definecolor{linkblue}{rgb}{0.21,0.49,0.74}
\usepackage[breaklinks,colorlinks,allcolors=linkblue]{hyperref}

\title{A Controlled Evaluation of Model Rankings and Input Reliance in Surface Water Segmentation}

\author{
  Kittipat Phunjanna$^{1,2}$, Kristóf Karacs$^{2}$, Chayut Ngamkhanong$^{3}$\\
  {\small $^{1}$Erasmus Mundus Joint Master in Image Processing and Computer Vision (IPCV),}\\
  {\small $^{2}$Pázmány Péter Catholic University}\\
  {\small $^{3}$Chulalongkorn University}\\
  {\small Preprint.}
}

\begin{document}
\maketitle

\begin{abstract}
Performance evaluation for surface-water segmentation commonly uses an aggregate metric such as global intersection-over-union (IoU) to rank model configurations. However, a configuration ranking does not by itself establish why one system performs better, whether a close ordering is stable, or how strongly predictions rely on individual inputs. We examine these distinctions primarily on Sen1Floods11 through repeated configuration comparisons, paired test-chip analysis, fixed-checkpoint input stress tests, and geographic reweighting, with a targeted secondary evaluation of supervised input configurations on GEOID-Flood. The cross-modal student achieves the highest three-seed mean IoU on Sen1Floods11, but close orderings vary across seeds and geographic weighting, while ancillary-input rankings differ between Swin-UNet and U-Net. The GEOID-Flood evaluation shows substantial agreement in supervised ancillary-input effects, although the exact architecture ordering remains configuration dependent. Fixed-checkpoint tests further establish reliance on terrain and WorldCover without establishing a clean-input performance benefit, while target semantics and the later WorldCover prior restrict the evaluation to retrospective all-water segmentation. These results show that aggregate metrics remain useful for ranking complete configurations, but ranking stability, component attribution, input reliance, and deployment scope require distinct evidence. Performance evaluation should therefore match the evidence reported to the claim being made.
\end{abstract}

\section{Introduction}
\label{sec:intro}

Synthetic Aperture Radar (SAR) observes floods through cloud and at night,
making Sentinel-1 useful when rapid optical mapping is unavailable
\cite{torres_sentinel1_2012,shen_inundation_sar_review_2019}. Yet low radar
return is not a water label: radar shadow and smooth dry surfaces can also
produce low backscatter \cite{shen_inundation_sar_review_2019}. Modern flood
mapping pipelines therefore combine SAR with terrain, hydrology, land cover,
weak labels, or optical information available only during training
\cite{bonafilia_sen1floods11_2020,paul_ganju_2021,garg_crossmodal_2023}.
Their benchmark score is the outcome of all these choices, not an explanation
of their individual contributions.

\begin{figure*}[t]
  \centering
  \includegraphics[width=0.98\textwidth]{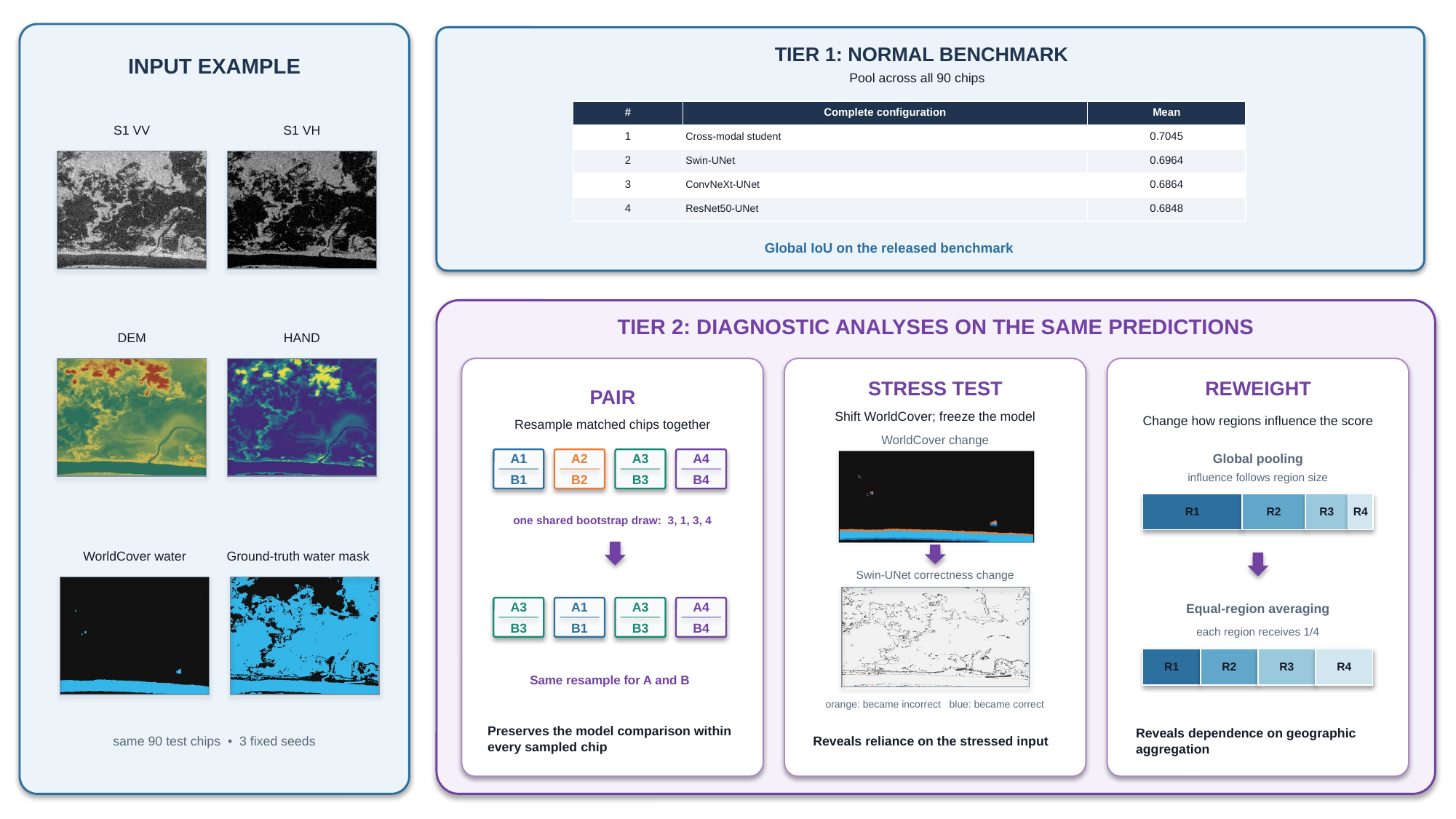}
  \caption{The same matched evidence supports different claims. Pooling ranks
  complete configurations; paired test-chip resampling diagnoses composition
  sensitivity; fixed-checkpoint input stress tests establish reliance under a
  declared input change; and geographic reweighting tests aggregation
  dependence. None alone identifies component causality or deployment
  performance.}
  \label{fig:method-flow}
\end{figure*}

Sen1Floods11 creates a particularly useful setting for this question
\cite{bonafilia_sen1floods11_2020}. Its released HandLabeled foreground is an
\emph{all-water} target on flood-event chips, combining permanent and
non-permanent water. Prior work improves this benchmark by bundling several
choices: the dataset supplies optical-index-derived weak labels, semi-supervised
methods exploit additional unlabeled or weakly labeled samples, and cross-modal
distillation combines a multimodal teacher with ancillary geophysical inputs
and a deployable SAR student
\cite{bonafilia_sen1floods11_2020,paul_ganju_2021,garg_crossmodal_2023}.
These results establish the value of complete pipelines, but do not by
themselves identify whether a gain comes from architecture, ancillary priors,
privileged supervision, or their interactions, nor whether a close ordering is
stable across training seeds and geographic composition. This gap motivates an
evaluation focused on attribution rather than another architecture.
Figure~\ref{fig:method-flow} summarizes the claims supported by each analysis.
The input examples include Sentinel-1 (S1) vertical transmit--vertical receive
(VV), vertical transmit--horizontal receive (VH), a digital elevation model
(DEM), and Height Above Nearest Drainage (HAND).

These distinctions matter for both architecture and ancillary inputs.
Comparing two end-to-end systems can show which complete configuration performs
better, but not which changed component caused the difference, because
initialization, parameterization, and optimization may also change
\cite{chen_closer_2019,bouthillier_variance_2021}. Likewise, training a model
with an ancillary input does not show that its predictions actually depend on
that input. Fixed-checkpoint input stress tests change one input while holding
the model fixed \cite{fong_extremal_2019}, while geographic reweighting tests
whether close rankings persist when different parts of the benchmark receive
different influence \cite{stehman_spatial_unit_2011,wang_finegrained_iou_2023}.

The evaluation addresses three research questions:
\begin{itemize}
  \item \emph{RQ1:} To what extent do close global-IoU rankings remain stable
  across the evaluated training seeds and test composition?
  \item \emph{RQ2:} Which conclusions about component attribution and input
  reliance are supported by configuration comparisons and fixed-checkpoint
  input stress tests?
  \item \emph{RQ3:} How do geographic reweighting, target semantics, and data
  provenance constrain the evaluation scope of the reported scores?
\end{itemize}
The evidence progresses from configuration ranking through ranking stability, component attribution, and input reliance to evaluation scope. The ranking remains valid for the evaluated configurations, but each additional analysis supports a different class of claim. The central principle is therefore to match the evaluation evidence to the claim being made, rather than treating an aggregate ranking as evidence for component or deployment conclusions.

\section{Related Work}
\label{sec:related}

\paragraph{SAR water mapping and ancillary information.}
Reviews of deep-learning flood mapping identify generalization to unseen cases
and uncertainty treatment as open problems \cite{bentivoglio_flood_review_2022}.
Sen1Floods11 provides geographically distributed Sentinel-1 imagery with
manual and automatically generated surface-water labels
\cite{bonafilia_sen1floods11_2020}. This study evaluates the released
HandLabeled all-water mask using the directly comparable 10\,m protocol
\cite{garg_crossmodal_2023}. U-Net and encoder--decoder descendants are common
segmentation baselines \cite{ronneberger_unet_2015,he_resnet_2016}, while
attention, convolutional, and transformer variants modify their
representations \cite{roy_scse_2018,liu_convnext_2022,cao_swinunet_2021}. Digital
Elevation Model (DEM) and Height Above Nearest Drainage (HAND) encode terrain
and drainage position \cite{nobre_hand_2011}; European Space Agency (ESA)
WorldCover supplies a global 10\,m land-cover map \cite{zanaga_esa_2022}. Such
inputs can help resolve SAR ambiguities \cite{shen_inundation_sar_review_2019},
but a high all-water IoU alone does not reveal whether predictions use
event-time SAR, recover static water aligned with a prior, or combine both.
Related Sen1Floods11 work separates permanent from temporary water
\cite{bai_permanent_temporary_water_2021}, while a direct benchmark comparison
reports qualitative limitations that complicate event-level interpretation
\cite{bereczky_sen1floods11_benchmark_2022}.

\paragraph{Training-only optical supervision.}
Cross-modal distillation can use a Sentinel-1/Sentinel-2 teacher to supervise a
deployable Sentinel-1 student that does not receive optical imagery at
inference \cite{garg_crossmodal_2023}. Related remote-sensing work studies
semi-supervised learning \cite{paul_ganju_2021}, multimodal pretraining
\cite{linial_mixmae_2025}, sensor fusion \cite{mcmillen_fuseform_2025}, and
larger multisensor corpora \cite{guo_skysense_2024}. Sen1Floods11 fusion studies
compare inference-time Sentinel-1, Sentinel-2, and elevation inputs
\cite{konapala_sentinel_diversity_2021}, whereas optical-teacher distillation
has also been used to train SAR-only segmentation students
\cite{nair_crossmodal_distillation_2024}. These regimes vary in training data
and optimization as well as supervision. The practical pipeline comparison
preserves those differences; the narrower fixed-teacher block holds teacher
outputs constant while varying deployable inputs. Sentinel-2 is used only
during training.

\paragraph{Evidence beyond aggregate scores.}
Benchmark re-evaluations test competing explanations of apparent rankings by
separating distribution shift from alternative causes
\cite{recht_imagenet_2019}, testing reliance on image cues
\cite{geirhos_texture_2019}, and controlling pipeline details to improve
method comparability \cite{chen_closer_2019}. Scores also vary with
initialization, sampling, and pipeline choices
\cite{bouthillier_variance_2021,reimers_score_distributions_2017}; benchmark
design guidance therefore emphasizes significance and replicability
\cite{reuel_betterbench_2024}. In flood mapping, high intra-dataset scores need
not transfer across datasets \cite{portales_crossdataset_flood_2025}.
Dataset-level IoU weights regions through their foreground unions
\cite{wang_finegrained_iou_2023}, while changing the spatial assessment unit
can change the summary \cite{stehman_spatial_unit_2011}, and fine-grained error
decompositions expose boundary, extent, and segment errors hidden by IoU
\cite{bernhard_beyond_iou_2024}. The present evaluation combines paired
test-chip analysis
\cite{dror_significance_2018}, matched input stress tests, and geographic
reweighting to distinguish a valid configuration ranking from unsupported
component and population claims.

\section{Evaluation Design}
\label{sec:protocol}

\subsection{Benchmark and evaluated configurations}

Experiments use Sen1Floods11 v1.1 at \(512\times512\) pixels and 10\,m ground
sampling.
The HandLabeled split used here contains 252 training, 89 validation, and 90 test
chips; a separate 15-chip Bolivia set is not part of that split. Each of
the ten named geographic groups occurs in train, validation, and test,
so they are composition strata rather than held-out domains. The WeaklyLabeled
pool contains 4,384 chips whose hard targets were generated from co-registered
Sentinel-2 indices \cite{bonafilia_sen1floods11_2020}.

The binary target is the released \texttt{LabelHand} all-water mask. Invalid
SAR pixels are ignored, probabilities are thresholded at 0.5, and the primary
outcome pools true positives (TP), false positives (FP), and false negatives
(FN) over all valid test pixels before computing foreground IoU. This matches
Garg et al.'s benchmark calculation \cite{garg_crossmodal_2023} and differs
from an equal-chip mean. The secondary metrics are the F$_1$ score, precision,
and recall.

The input channels are vertical transmit--vertical receive (VV), vertical
transmit--horizontal receive (VH), a digital elevation model (DEM), Height
Above Nearest Drainage (HAND), and a binary permanent-water prior (Water). DEM
records absolute ground elevation; HAND records elevation relative to connected
drainage and therefore describes local drainage position \cite{nobre_hand_2011}.
Both originate from the nominal 30\,m Copernicus GLO-30 DEM, with HAND derived
using eight-direction (D8) flow routing. Bilinear alignment to the model's
10\,m grid does not
create new 10\,m terrain detail. For
comparability with the benchmark SAR setup \cite{garg_crossmodal_2023}, VV and
VH backscatter values are clipped to \([-30,0]\) dB to bound extreme returns
and then linearly mapped to \([0,1]\). DEM is clipped to \([-50,3000]\) m and
HAND to \([0,100]\) m before the same scaling. Categorical WorldCover is
nearest-neighbor resampled to preserve its class codes, and class 80 becomes
Water \cite{zanaga_esa_2022}. WorldCover v200 represents 2021
\cite{zanaga_esa_2022}, whereas the evaluated SAR acquisitions span 2016--2019
\cite{bonafilia_sen1floods11_2020}. Consequently, this evaluation treats Water
as a retrospective static prior with temporal look-ahead, not as a
contemporaneous operational input.

Within the standard supervised training recipe, two complementary comparisons are evaluated. Hereafter, \emph{Swin-UNet} refers to the Swin-UNet configuration trained with this recipe; the teacher--student pipeline is identified separately as the \emph{cross-modal student}. First, six complete architecture configurations are compared under the same five-channel input recipe: vanilla U-Net, U-Net with a ResNet34 encoder and concurrent spatial and channel squeeze-and-excitation (scSE) attention, ResNet34-UNet, ResNet50-UNet, ConvNeXt-UNet, and Swin-UNet \cite{ronneberger_unet_2015,roy_scse_2018,he_resnet_2016,liu_convnext_2022,cao_swinunet_2021}. Four configurations contain 31.0--32.5M parameters; the two ResNet34 variants contain 24.4M and 24.6M. Only vanilla U-Net uses random initialization; the other five use ImageNet initialization. Accordingly, this experiment compares complete configurations with broadly comparable parameter counts rather than isolating architecture.

Second, ancillary-input configurations are compared for Swin-UNet and vanilla U-Net. Each model uses all eight subsets formed by VV+VH plus any combination of DEM, HAND, and Water. Because changing the subset also changes the input stem, these runs compare complete input configurations rather than isolate the effect of a single channel. Each configuration is trained with seeds 42, 1337, and 2026, balancing repeated training against coverage of the complete matrix within the available computational budget \cite{lakens_sample_size_2022,saha_edaps_2023,kumar_seabird_2024,herzog_olmoearth_2026}. Full training configurations and model parameter counts are provided in the supplementary material.

As a secondary evaluation, the U-Net/Swin-UNet input matrix is repeated on GEOID-Flood using post-event VV/VH and the same ancillary inputs \cite{chiriaco_geoid_flood_2026}, with permanent and flooded labels merged into an all-water target. The 100-epoch cap makes the design matched rather than identical; preprocessing and paired source-tile analyses are detailed in the supplement.

\subsection{Cross-modal and fixed-teacher comparisons}
\label{sec:gargmethod}

The cross-modal pipeline follows the teacher--student design of Garg
et al.~\cite{garg_crossmodal_2023}. A nine-channel Swin-UNet teacher receives
the five inputs and Sentinel-2 blue (B2), green (B3), red (B4), and
near-infrared (B8) bands. Trained on the 252 HandLabeled chips, it supplies
frozen soft targets for paired HandLabeled and WeaklyLabeled samples. The
ImageNet-initialized student uses only its declared inputs; hard
optical-index weak labels are replaced by teacher probabilities. Supervised
training uses no augmentation, a \(10^{-3}\) learning rate, and an early-stopped
200-epoch cap. Cross-modal teacher and student training uses flips and
right-angle rotations, a \(3\times10^{-4}\) learning rate, and a 100-epoch cap.
It also adds a teacher-training stage and replaces weak hard-label loss with
soft-target loss. Because the two pipelines also differ in augmentation, learning rate, epoch cap, and teacher training, their comparison evaluates complete training pipelines rather than isolating the effect of supervision. 


The primary cross-modal student uses all five non-optical inputs. A separate end-to-end no-Water pipeline retrains both teacher and student without WorldCover. For the fixed-teacher analysis, six student configurations within each seed share the same full-input teacher cache. This removes teacher-output variation while retaining differences in student input width and training trajectory.

\subsection{Fixed-checkpoint tests of input reliance}

Receiving an input and relying on it are distinct. The WorldCover stress test holds the checkpoint, SAR, other inputs, labels, and test chips fixed while changing only Water: the prior is zeroed, circularly translated by a random nonzero offset, or replaced by another chip’s mask. A separate localization test translates the original mask by 10, 30, 50, or 100 m in a deterministic chip-specific cardinal direction with nearest-edge extension. Water interior, boundary, and exterior strata are computed once from the clean prior. This identifies whether prediction changes are concentrated where the prior itself changes. Because translating a binary mask changes values mainly near its boundaries, such concentration does not by itself indicate a learned boundary-specific mechanism.


The terrain stress test similarly changes DEM and HAND for the five-channel cross-modal, Swin-UNet, and U-Net checkpoints at all three seeds, together with the three Swin-UNet VV+VH+HAND checkpoints. Each terrain input is replaced by a constant train-split median, translated locally by 10–100 m, replaced by a deterministic cross-chip donor surface, or modified within a local clipped patch by \(\pm\) one interquartile range (IQR). These tests measure the sensitivity of fixed checkpoints to specific out-of-distribution changes; they do not estimate clean-input benefits or naturally occurring failure rates. A 30 m translation corresponds to three pixels on the aligned 10 m model grid, not to the source resolution of the terrain data.

\subsection{Paired comparisons and test-composition sensitivity}
\label{sec:evaluation}

Let \(i\) index chips, \(m\in\mathcal M\) configurations, and
\(s\in\{42,1337,2026\}\) evaluated seeds. For a multiset \(\mathcal I\) of chip
indices, the pooled IoU of configuration \(m\) at seed \(s\) is
\begin{equation}
J_{ms}(\mathcal I)=
\frac{\sum_{i\in\mathcal I}\mathrm{TP}_{ims}}
{\sum_{i\in\mathcal I}
(\mathrm{TP}_{ims}+\mathrm{FP}_{ims}+\mathrm{FN}_{ims})}.
\label{eq:iou}
\end{equation}
For configurations \(A,B\in\mathcal M\), \(D_s(A,B)\) is their seed-specific
paired IoU difference in percentage points, and \(\bar D(A,B)\) averages that
difference over the three evaluated seeds:
\begin{equation}
\begin{aligned}
D_s(A,B)&=100[J_{As}(\mathcal I)-J_{Bs}(\mathcal I)],\\
\bar D(A,B)&=\frac{1}{3}\sum_s D_s(A,B).
\end{aligned}
\label{eq:paired}
\end{equation}
Thus \(D_s\) preserves seed-specific direction, whereas \(\bar D\) is an
arithmetic summary of three fixed runs, not inference to a population of
possible retraining runs.

Close pooled-IoU differences may depend on which test chips compose the
benchmark. To measure that sensitivity while preserving model pairing, each
bootstrap replicate samples 90 chip indices with replacement and applies the
same multiset to both models \cite{efron_tibshirani_1994}. For \(\bar D\), the
multiset is also shared across seeds. IoU is recomputed from pooled chip counts,
not pixel-level resampling, to retain within-chip spatial dependence. The
reported \(q_{.025},q_{.975}\) therefore describe sensitivity to changes in
the observed test composition; they do not estimate retraining variability.

Before using these ranges for categorical claims, candidate percentile,
bias-corrected and accelerated (BCa), studentized, and max-\(|t|\) intervals
were checked on empirical known-null and known-effect cases. Their calibration
did not support labels of separation,
equality, or equivalence. The validation and Holm-adjusted randomization
analysis is supplementary; the main analysis instead uses effect magnitude,
cross-seed direction, and matched input stress tests.

Using the same \(m\) and \(s\), let \(c\) index an input stress condition
applied to a fixed checkpoint, with \(c=0\) denoting its clean-input baseline.
The stress effect is
\begin{equation}
R_{msc}=100[J_{msc}(\mathcal I)-J_{ms0}(\mathcal I)],
\label{eq:stress}
\end{equation}
Three-seed stress summaries average \(R_{msc}\) within each common-chip
replicate. Main comparison and stress summaries use 50,000 resamples to
reduce Monte Carlo variation in the reported quantiles
\cite{efron_tibshirani_1994}. This increases numerical precision but not the
amount of independent test information; all summaries condition on the
evaluated checkpoints and fixed test population.

\subsection{Geographic composition sensitivity}

Global IoU gives greater influence to regions with larger foreground unions.
Counts are therefore also pooled within each named group \(g\), and the ten
group-level IoUs receive equal weight:
\begin{equation}
J^{\mathrm{macro}}_{ms}=\frac{1}{10}\sum_{g=1}^{10}J_{ms}(\mathcal I_g).
\label{eq:macro-iou}
\end{equation}
Region-cluster resampling and leave-one-region-out recomputation diagnose
dependence on the released composition. Because groups contain only 4--14
chips, occur in every split, and are not verified event identities, these are
not estimates of new-country or new-event performance
\cite{wang_finegrained_iou_2023,stehman_spatial_unit_2011}.

\section{Results}
\label{sec:results}

\subsection{Configuration ranking and ranking stability}

\begin{figure*}[t]
  \centering
  \includegraphics[width=0.98\textwidth]{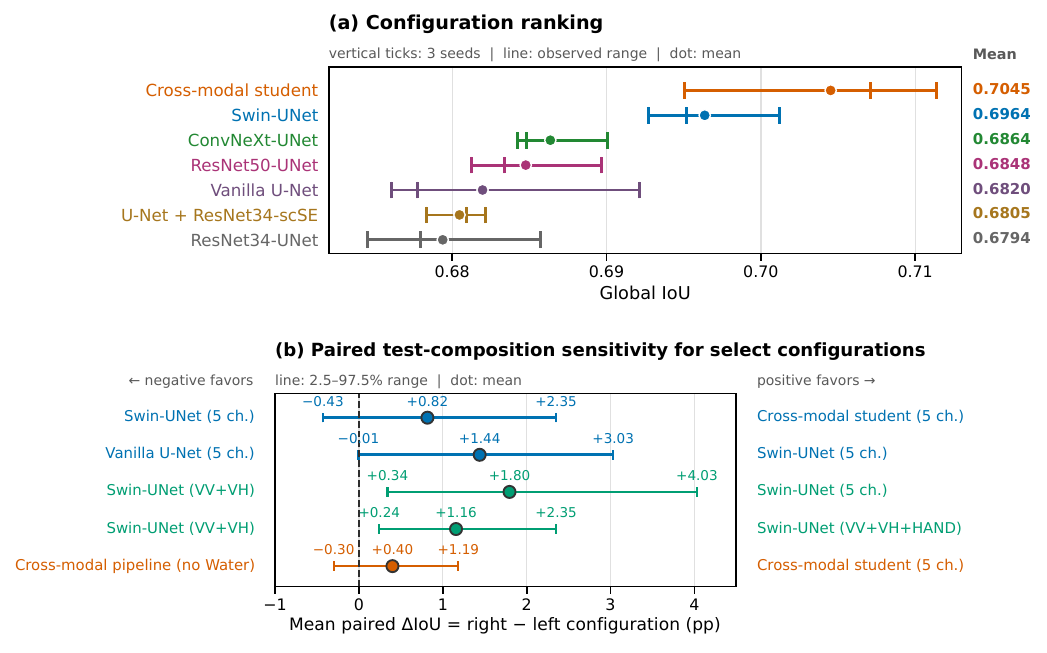}
  \caption{Configuration ranking and paired test-composition sensitivity.
  (a) Three-seed means and observed ranges. (b) Select comparisons grouped by
  color: matched five-channel inputs (blue), matched supervised Swin-UNet
  architecture/training with varied inputs (green), and cross-modal pipelines
  with/without Water (orange). Differences average paired chip resamples across
  seeds; positive favors right. Bars are descriptive 2.5th--97.5th ranges
  conditional on fixed seeds. Five channels are VV, VH, DEM, HAND, and Water.}
  \label{fig:ranking-paired}
\end{figure*}

Figure~\ref{fig:ranking-paired} begins with the conventional configuration
ranking. The cross-modal student has the highest three-seed mean, 0.7045, and
the seed-42 IoU of 0.7114 is the best individual run. Swin-UNet is the highest
supervised configuration at every seed, with mean 0.6964. The deterministic
VH-Otsu \texttt{S1OtsuLabelHand} rasters distributed with Sen1Floods11
\cite{bonafilia_sen1floods11_2020} reach 0.5458 IoU on the same test chips.
The learned configurations therefore remain clearly separated from this
training-free reference.

The close cross-modal--Swin-UNet ordering is less stable. Relative to
Swin-UNet, the cross-modal student's IoU changes by (+1.87), (-0.01), and
(+0.59) percentage points across the three seeds. The descriptive paired
test-chip resampling range spans zero at every seed. The mean ranking identifies
the cross-modal student as the leading candidate on the archived runs, while
the seed-specific results do not establish the same ordering after retraining.

Geographic reweighting produces the same interpretive boundary. At seed 1337,
pooled IoU ranks Swin-UNet, cross-modal, full U-Net, then two-channel U-Net;
equal-region IoU instead ranks full U-Net, Swin-UNet, two-channel U-Net, then
cross-modal. Cross-modal--Swin-UNet test-chip and region-cluster ranges span zero at
every seed, and leave-one-region-out recomputation crosses zero in two rows.
The average lead therefore ranks the released benchmark composition without
establishing the ordering for another geographic population.

\subsection{Configuration comparisons and component attribution}

The supervised architecture comparison ranks complete configurations rather
than isolated architectural changes. The six five-channel mean IoUs occupy a
1.70-point band. Swin-UNet changes IoU relative to vanilla U-Net by (+1.67),
(+0.30), and (+2.35) points across seeds, but the descriptive paired ranges
span zero. Initialization, parameterization, encoder design, and optimization
also differ across rows. Swin-UNet is therefore the highest-ranked member of the
narrow controlled architecture group, not evidence of an isolated transformer
advantage.

The complete Swin-UNet and U-Net ancillary-input matrices likewise do not support a
universal channel prescription. Full input has the highest Swin-UNet mean, 1.80
points above VV+VH, while adding Water alone gives the highest U-Net mean, 1.12
points above the U-Net VV+VH reference. For U-Net, the full five-channel configuration is lower than VV+VH+Water.
VV+VH+HAND is the most consistent Swin-UNet candidate, with a positive difference
from VV+VH at all three seeds, but the same consistency is not reproduced by
U-Net. The result supports an architecture-dependent configuration rather than
a universal HAND effect.

The cross-modal comparison also supports a complete training pipeline rather
than an isolated training-only optical supervision effect. Both the
cross-modal pipeline and Swin-UNet use optical-derived training targets,
and the recipes differ beyond target softness. Relative to retraining the
teacher and student without Water, the full pipeline changes direction across
seeds. Holding teacher outputs fixed removes teacher-output variation, but
student input width and training trajectory still change, and no stable
student-input leader emerges. DEM+Water has the highest observed fixed-teacher
mean, followed by DEM and Water; the complete seed results are reported in the
supplementary material.

Together, the architecture, ancillary-input, and cross-modal comparisons rank
the evaluated complete configurations. Component attribution requires a more
tightly matched comparison than the available configuration rows provide.

\begin{table}[t]
\centering\footnotesize
\caption{Cross-dataset ancillary-input comparisons. Each cell lists the Sen1Floods11/GEOID-Flood change in three-seed mean global IoU (pp) from the corresponding VV+VH baseline. Seven subsets for each architecture yield 14 comparisons; $\dagger$ marks the three direction disagreements.}
\label{tab:geoid-cross-dataset}
\setlength{\tabcolsep}{2.5pt}
\begin{tabular}{lrr}
\toprule
Ancillaries added & U-Net $\Delta$IoU & Swin-UNet $\Delta$IoU \\
to VV+VH & S11 / GEOID-Flood & S11 / GEOID-Flood \\
\midrule
DEM & -0.20 / -5.68 & -0.06 / -1.34 \\
HAND & +0.47 / +0.02 & +1.16 / +2.49 \\
Water & +1.12 / +2.50 & -0.24 / +3.04\textsuperscript{$\dagger$} \\
DEM+HAND & +0.61 / -1.17\textsuperscript{$\dagger$} & +1.24 / +0.33 \\
DEM+Water & -0.01 / +3.42\textsuperscript{$\dagger$} & +0.33 / +3.24 \\
HAND+Water & +0.66 / +3.14 & +1.24 / +4.12 \\
DEM+HAND+Water & +0.15 / +3.48 & +1.80 / +3.57 \\
\bottomrule
\end{tabular}
\end{table}

\paragraph{Secondary evaluation on GEOID-Flood.}
Eleven of the 14 changes in Table~\ref{tab:geoid-cross-dataset} have the same
direction on Sen1Floods11 and GEOID-Flood. Full-input Swin-UNet exceeds
full-input U-Net by 1.19 percentage points, with a descriptive paired
source-tile resampling range of $[+0.90,+1.50]$ points; however, the VV+VH
architecture difference changes sign across seeds. This evaluation does not
cover cross-modal supervision, fixed-checkpoint input reliance, or unseen-event
performance.

\subsection{Fixed-checkpoint input reliance}

Fixed-checkpoint input stress tests address a different question from the
clean-input configuration comparisons. Table~\ref{tab:terrain-stress} reports
the mean ancillary-channel effects. Cross-chip HAND donor substitution
reduces mean IoU by 8.67, 8.11, and 5.03 points for the five-channel
cross-modal, Swin-UNet, and U-Net checkpoints; the corresponding DEM losses are
2.17, 1.15, and 3.06 points. For Swin-UNet VV+VH+HAND, HAND donor substitution
causes a 17.24-point loss. Median replacement produces smaller HAND losses,
while 100\,m terrain translations produce much smaller changes. Fixed
predictions therefore depend on chip-assigned terrain under the declared input
changes, especially HAND, without establishing a clean-input performance
benefit.

\begin{table*}[t]
\centering\small
\caption{Inference-time ancillary-channel stress averaged over three seeds. Entries are $\Delta$IoU (stress $-$ clean), in percentage points. The final column uses Swin-UNet VV+VH+HAND; all others use five-channel checkpoints. Translations move the complete ancillary surface (100 m is 10 model-grid pixels). Water's median and interquartile range (IQR) are zero, so median replacement equals zeroing and the $\pm$IQR patch is omitted. Paired descriptive ranges are reported in the supplement.}
\label{tab:terrain-stress}
\setlength{\tabcolsep}{2.5pt}
\begin{tabular}{lrrrrrrrrrr}
\toprule
& \multicolumn{3}{c}{Cross-modal} & \multicolumn{3}{c}{Swin-UNet} & \multicolumn{3}{c}{U-Net} & Swin-UNet 3-ch. \\
\cmidrule(lr){2-4}\cmidrule(lr){5-7}\cmidrule(lr){8-10}
Stress condition & DEM & HAND & Water & DEM & HAND & Water & DEM & HAND & Water & HAND \\
\midrule
constant median (zero for Water) & -0.86 & -4.00 & -1.85 & -0.36 & -2.33 & -0.80 & -1.31 & -0.81 & -2.27 & -4.87 \\
edge shift 10 m (1 px) & +0.00 & -0.08 & -0.04 & +0.00 & +0.00 & -0.02 & -0.00 & -0.01 & -0.03 & +0.00 \\
edge shift 30 m (3 px) & +0.00 & -0.03 & -0.26 & -0.00 & -0.06 & -0.13 & -0.00 & -0.06 & -0.21 & -0.06 \\
edge shift 50 m (5 px) & +0.00 & -0.05 & -0.48 & -0.00 & -0.17 & -0.29 & -0.00 & -0.12 & -0.41 & -0.18 \\
edge shift 100 m (10 px) & +0.01 & -0.32 & -0.67 & -0.00 & -0.41 & -0.50 & -0.01 & -0.25 & -0.70 & -0.45 \\
donor substitution & -2.17 & -8.67 & -2.98 & -1.15 & -8.11 & -1.65 & -3.06 & -5.03 & -3.11 & -17.24 \\
local $100{\times}100$ patch $\pm$IQR & -0.17 & -0.62 & --- & -0.11 & -0.38 & --- & -0.14 & -0.21 & --- & -0.57 \\
\bottomrule
\end{tabular}
\end{table*}

WorldCover has a strong relationship with the released target. Water occupies
6.50\% of valid pixels but contains 46.0\% of reference-water positives. All
27 combinations of three stress conditions, three model families, and three
seeds reduce IoU. Donor mask substitution produces the largest mean loss in
each family: 1.65 points for Swin-UNet, 2.98 for the cross-modal student, and 3.11
for U-Net. Because only the declared input changes at a fixed checkpoint, the
result establishes WorldCover reliance under the tested conditions. The result
does not distinguish copying from contextual use or establish a clean-input
benefit.

\begin{figure}[t]
  \centering
  \includegraphics[width=\columnwidth]{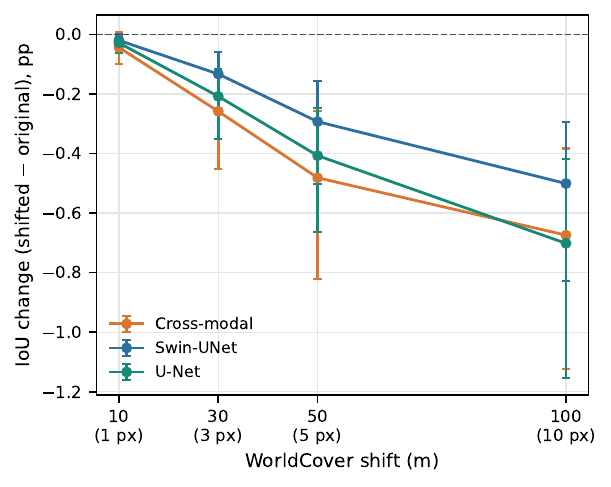}
  \caption{Effect of inference-time WorldCover shifts. Curves show three-seed
  mean $\Delta$IoU (shifted $-$ original), with descriptive paired quantiles;
  checkpoints and all other inputs remain fixed.}
  \label{fig:worldcover-shift-curve}
\end{figure}

\begin{figure*}[t]
  \centering
  \includegraphics[width=0.98\textwidth]{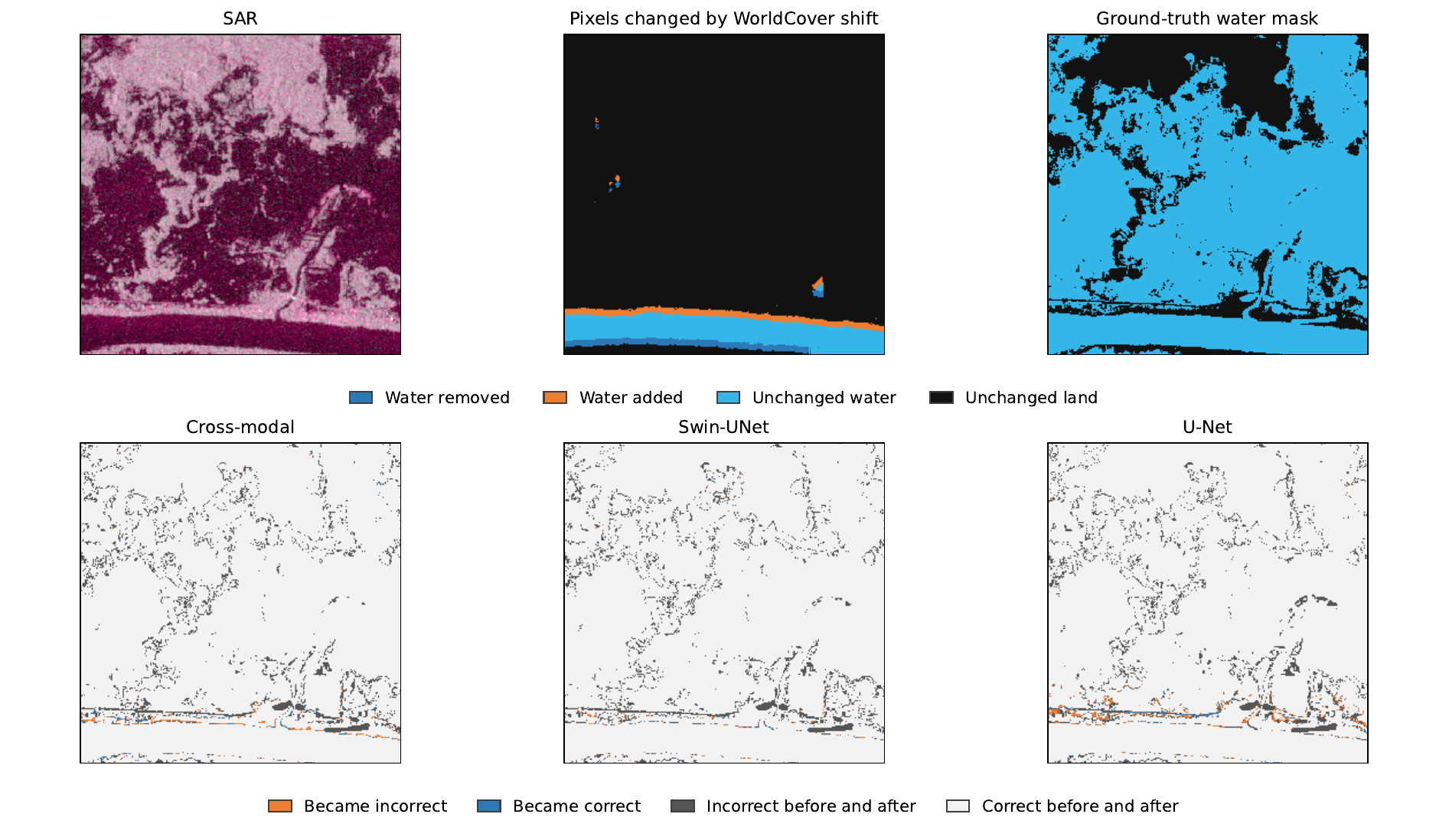}
  \caption{Prediction-correctness changes after a 100\,m WorldCover shift.
  Top: SAR, changed Water pixels, and ground truth; bottom: three model
  families. The large shift aids visibility; Figure~\ref{fig:worldcover-shift-curve}
  reports all distances.}
  \label{fig:worldcover-correctness}
\end{figure*}

Figure~\ref{fig:worldcover-shift-curve} shows that the global IoU loss increases
with the WorldCover shift distance for all three model families.
Figure~\ref{fig:worldcover-correctness} shows where the 100\,m shift changes
WorldCover and prediction correctness in a representative chip. Across all
test chips at 30\,m, the shift changes Water on only 0.64\% of valid pixels.
Prediction-flip rates on this changed support are 10.74\%, 13.24\%, and
14.66\% for cross-modal, Swin-UNet, and U-Net, compared with at most 0.12\%
elsewhere. Prediction changes are therefore concentrated where WorldCover
changes. However, shifting a binary mask changes values mainly near its
boundary, so this concentration does not establish a learned boundary-specific
mechanism.

\subsection{Evaluation scope}

The input-reliance results must be interpreted together with target semantics
and data provenance. The released target combines permanent and non-permanent
water, so global IoU rewards recovery of both. WorldCover represents 2021,
whereas the evaluated SAR acquisitions span 2016--2019. WorldCover is therefore
a retrospective static prior with temporal look-ahead rather than a
contemporaneous operational input.

The geographic analysis provides a second scope boundary. The ten named groups
occur in every split, contain few test chips, and are not verified event
identities. Geographic reweighting measures dependence on the released test
composition, not performance in an unseen event, country, sensor condition, or
acquisition geometry.

The global-IoU ranking remains valid for the evaluated complete configurations
on the released test split. The supported interpretation is narrower than a
component explanation or deployment claim: the results characterize
retrospective all-water segmentation under the archived training runs and
released benchmark composition.

\section{Discussion, Implications, and Limitations}
\label{sec:discussion}

\paragraph{Interpretation of the ranking.}
%

The evaluation does not invalidate the global-IoU ranking; it narrows its interpretation. The cross-modal student’s highest three-seed mean identifies a candidate system for further validation, while Swin-UNet remains a credible alternative. However, the close difference does not isolate training-only optical supervision because the pipelines also differ in other training choices. Likewise, the architecture and ancillary-input comparisons rank complete configurations rather than individual components, and the different channel rankings for Swin-UNet and U-Net prevent a universal channel prescription. A high score therefore supports selection of a tested configuration, not attribution of its performance to one architectural or input component. 
Global IoU can also conceal boundary, extent, or segment errors \cite{bernhard_beyond_iou_2024}; prediction agreement and representation similarity were not measured.

\paragraph{Configuration performance and input reliance.}
%

Clean-input configuration comparisons and fixed-checkpoint stress tests answer different questions: the former identify useful input configurations under the tested training recipes, whereas the latter establish that predictions depend on terrain and WorldCover under the declared input changes. Such dependence does not establish a clean-input performance benefit or a naturally occurring failure rate. Practically, static ancillary inputs should therefore be treated as system dependencies, and validation should consider missing, stale, misregistered, or incorrectly assigned inputs when relevant to the intended operating population.

\paragraph{Target semantics and data provenance.}
%

Because the target combines permanent and event-specific water and WorldCover postdates the evaluated SAR acquisitions, the reported scores describe retrospective all-water segmentation with temporal look-ahead rather than transient-inundation or deployment performance. The WorldCover stress tests establish prediction responses to the declared input changes, but do not distinguish beneficial context, copying, input-stem interactions, or brittle reliance.

\paragraph{Uncertainty and population scope.}
%

Three seeds reveal changes in close orderings but do not characterize the full distribution of retraining outcomes. Paired test-chip resampling measures sensitivity to the observed test composition at fixed checkpoints, not optimization uncertainty. The reported ranges remain descriptive because the candidate interval procedures did not pass every empirical coverage check; spanning zero establishes neither equality nor equivalence.

\paragraph{Implications for benchmark evaluation.}
Global IoU ranks configurations; repeated and paired analyses assess stability;
fixed-checkpoint stresses assess input reliance; and target semantics,
provenance, and geographic reweighting define scope. This separation avoids
unsupported component or deployment interpretations
\cite{recht_imagenet_2019,reimers_score_distributions_2017,
reuel_betterbench_2024}.

Future evaluation should separate transient and permanent water, use temporally
valid priors, match training choices in component comparisons, test more seeds on held-out events, and specify
intended-use metrics, calibration, and costs. The supplement details complete configuration matrices, GEOID-Flood preprocessing, and source-tile and activation-label composition-sensitivity analyses.


\section{Conclusion}
\label{sec:conclusion}

Performance evaluation should carefully distinguish between the construct captured by a benchmark metric and the substantive claims subsequently inferred from the resulting rankings. In this study, global IoU measures overlap with the Sen1Floods11 all-water target; a secondary evaluation on GEOID-Flood repeats the supervised ancillary-input comparisons. The cross-modal student has the highest Sen1Floods11 three-seed mean, but this ranking alone does not establish stability across retraining or test composition, isolate architecture or input contributions, or demonstrate input reliance. 
On GEOID-Flood, most supervised ancillary-input changes agree in direction, but cross-modal, reliance, and deployment claims are not evaluated.

The broader lesson is to match evidence to claims.
Aggregate metrics rank systems; paired analyses test stability; interventions test reliance; semantics, provenance, and geography define scope.
A higher score aids model selection but neither explains why a system performs better nor establishes performance in another setting.

{
    \small

}

\end{document}